\documentclass[lettersize, conference]{IEEEtran}
\IEEEoverridecommandlockouts
\usepackage{cite}
\usepackage{amsmath,amssymb,amsfonts}
\usepackage{algorithmic}
\usepackage{graphicx}
\usepackage{textcomp}
\usepackage{xcolor}
\usepackage{tabularx}
\usepackage{graphicx}
\usepackage{dblfloatfix}
\usepackage{hyperref}

\usepackage{enumitem}
\usepackage[font=normalsize,labelfont=bf]{caption}
\def\BibTeX{{\rm B\kern-.05em{\sc i\kern-.025em b}\kern-.08em
    T\kern-.1667em\lower.7ex\hbox{E}\kern-.125emX}}

\begin{document}
\title{\LARGE \bf 
A Co-Design Framework for High-Performance Jumping of a Five-Bar Monoped with Actuator Optimization \\}


\author{Aastha Mishra$^{1}$, Aman Singh$^{1}$, and Shishir Kolathaya$^{1}$%
\thanks{This research is supported by the AI \& Robotics Technology Park (ARTPARK) at IISc.}%
\thanks{$^{1}$Department of Cyber-Physical Systems (CPS), Indian Institute of Science (IISc), Bengaluru, India.
{\tt\scriptsize \{aasthamishra, saman, shishirk\}@iisc.ac.in}}%
}%













\maketitle
\thispagestyle{empty}
\pagestyle{empty}

\begin{abstract} 
The performance of legged robots depends strongly on both mechanical design and control, motivating co-design approaches that jointly optimize these parameters. However, most existing co-design studies focus on link dimensions and transmission ratios while neglecting detailed actuator design, particularly motor and gearbox parameter optimization, and are largely limited to serial open-chain mechanisms. In this work, we present a co-design framework for a planar closed-chain five-bar monoped that jointly optimizes mechanical design, motor and gearbox parameters, and control parameters for dynamic jumping. The objective is to maximize jump distance while minimizing mechanical energy consumption. The framework employs a two-stage optimization approach, where actuator optimization generates a mapping from gear ratio to actuator mass, efficiency, and peak torque, which is then incorporated into CMA-ES-based co-design optimization of the robot design and control parameters. Simulation results demonstrate an improvement of approximately 30.4\% in jump distance and an 11.5\% reduction in mechanical energy consumption compared to a nominal design, highlighting the effectiveness of the proposed framework for high-performance and energy-efficient planar jumping.
\end{abstract}



\section{Introduction}

Legged robots are important for navigating uneven and complex terrains where wheeled or tracked systems cannot operate effectively. In addition to walking and running, behaviors such as jumping allow legged robots to overcome obstacles and terrain gaps. The performance of these robots during such dynamic tasks depends jointly on their mechanical design and the control policies used to actuate them \cite{song2025codesign}.  Additionally, power efficiency while performing such tasks, remains crucial, as minimizing energy use during jumps is essential for deployment. 

Recent research has explored frameworks that jointly consider mechanical design, and control parameters within a unified optimization framework, commonly referred to as co-design. Existing co-design studies such as \cite{StarlETHCooptimization,MetaRLCodesign} optimize parameters including link lengths and transmission ratios for locomotion tasks such as walking. Similarly, \cite{PantherLeg} focuses on optimizing the motor and gearbox pair to improve jump height but does not consider energy consumption or variations in link lengths. 
Several existing works \cite{MetaRLCodesign, StarlETHCooptimization, Co-designing_versatile_quadruped_robots_for_dynamic_and_energy-efficient_motions} optimize structural parameters such as link lengths and transmission ratios, but do not explicitly incorporate detailed planetary gearbox design, while \cite{JointOptIFT} optimizes link lengths and actuator attachment points without considering gearbox parameters. Studies such as \cite{ Simulation_Aided_Co-Design_for_Robust_Robot_Optimization} include models of motor and gearbox friction but focus primarily on belt-driven transmissions rather than planetary gear systems. More recent works such as \cite{singh2025compact, KaistHound} optimize gearbox parameters for actuator design; however, they do not simultaneously consider controller optimization, motor selection, or the structural parameters of legged robots. A recent work \cite{singh2025co} incorporates gearbox optimization within a co-design framework but is limited to serial 2R (2-Revolute) mechanisms and does not optimize motor selection, while \cite{de2021control} compares different leg configurations without optimizing actuators.  

In the discussed literature, most works do not perform joint optimization of motor and gearbox parameters. Additionally, with the exception of \cite{de2021control}, most studies focus on planar serial open-chain mechanisms rather than closed-chain mechanisms, despite several works demonstrating the advantages of parallel closed-chain designs. For example, parallel mechanisms can provide higher structural stiffness compared to serial counterparts \cite{pandilov2014comparison}. Similarly, \cite{koutsoukis2021effect} demonstrates energy reduction using a five-bar linkage with coaxially actuated joints. Furthermore, \cite{kenneally2016design} shows that five-bar mechanisms can achieve improved foot force production compared to serial 2R legs. The five-bar mechanism is particularly attractive because the actuators can be mounted on the robot base, reducing distal mass and enabling highly dynamic locomotion. These advantages make 
five-bar mechanisms a promising choice for dynamic legged robots.

The existing co-design literature primarily focuses on optimizing link lengths and transmission ratios, while often neglecting detailed motor and gearbox parameter optimization. Furthermore, most prior works consider serial open-chain mechanisms, with limited attention given to parallel closed-chain mechanisms such as five-bar linkages. To address these limitations, this paper proposes a co-design framework for a closed-chain parallel planar five-bar mechanism for dynamic jumping tasks. The framework jointly optimizes the mechanical design, motor, gearbox parameters, and control parameters to improve performance. The main contributions of this work are as follows:

\begin{itemize}
    \item We propose a novel co-design optimization methodology to optimize design and control parameters of a closed-chain planar five-bar mechanism to maximize jump distance while minimizing mechanical energy consumption.
    \item We integrate motor and detailed gearbox optimization within the co-design framework. To the best of our knowledge, this is the first work to incorporate motor and detailed gearbox parameter optimization within a co-design framework for a closed-chain mechanism, specifically a five-bar mechanism.
\end{itemize}

\section{Preliminaries}\label{leg_design}
In this section, we describe the symmetrical five-bar mechanism, the leg-link mass model, the considered gearbox architectures, and the control framework used for jumping. The symmetrical five-bar leg mechanism is optimized using the proposed methodology in Section \ref{opt_framework}.

\begin{figure}[ht]
\centering
\includegraphics[width=0.4\linewidth]{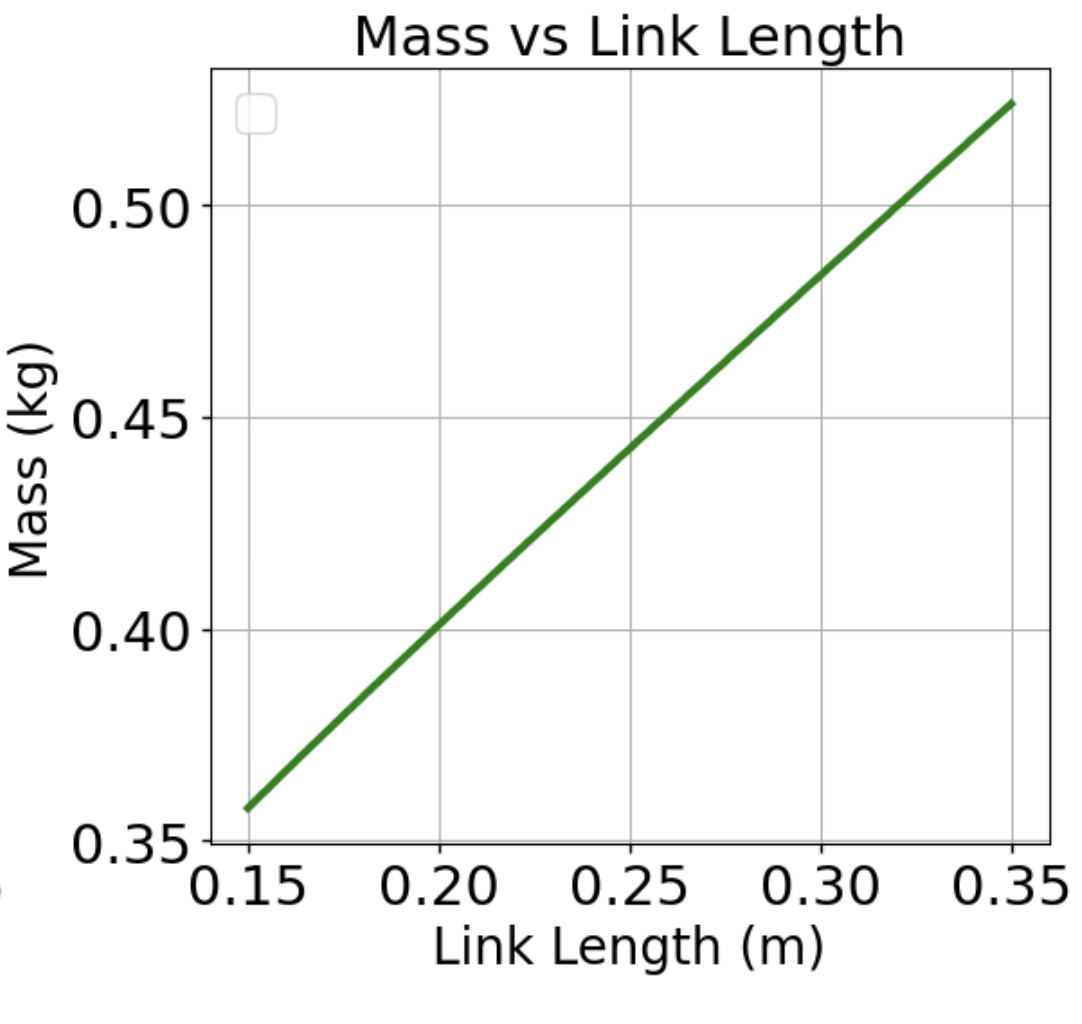}
\vspace{-1em}
\caption{Link-length to mass mapping, derived from the leg link mass model described in Section~\ref{five bar arch}.}
\label{fig:mass_v_ll}
\vspace{-1.0em}
\end{figure}

\subsection{Design of the Five-Bar Monoped}
\label{five bar arch}

The monoped employs a 2-DOF planar parallel five-bar leg mechanism with two actuated hip joints separated by a finite distance. This actuator placement reduces leg inertia and improves hopping performance.

\begin{figure}
\centering
\includegraphics[width=0.8\linewidth]{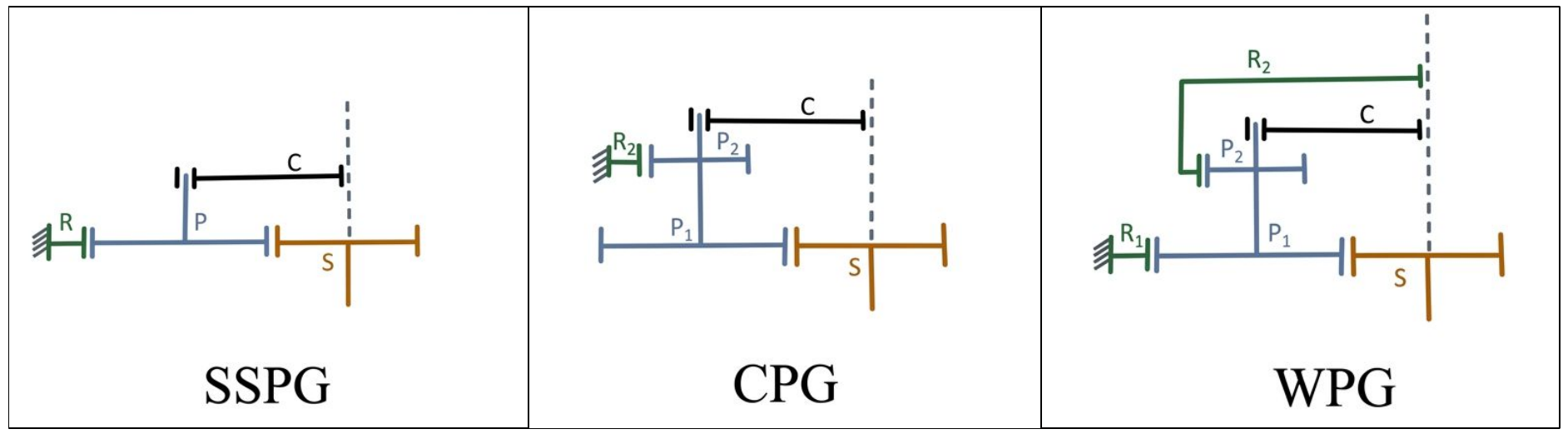}
\caption{Line diagrams of gearbox configurations. \textbf{SSPG:} Sun ($S$) input, Ring ($R$) fixed, Carrier ($C$) output. \textbf{CPG:} Sun ($S$) input, Ring ($R_2$) fixed, Carrier ($C$) output; planets ($P_1, P_2$) rigidly attached. \textbf{WPG (3K):} Sun ($S$) input, Ring ($R_1$) fixed, Ring ($R_2$) output.}
\label{fig:line_diagram}
\end{figure}

Each actuator consists of a brushless DC (BLDC) motor coupled with a planetary gearbox. Three planetary gearbox architectures are considered: Single-Stage Planetary Gearbox (SSPG), Compound Planetary Gearbox (CPG), and Wolfrom Planetary Gearbox (WPG). The SSPG uses a fixed ring gear with the carrier as output, while the CPG and WPG employ rigidly connected planet gears to achieve higher reduction ratios. The WPG (3K configuration)~\cite{3KgearOpt} additionally uses two concentric ring gears, enabling very high gear ratios within a compact volume. Expressions for gear ratios and efficiencies are provided in~\cite{singh2025compact}. Motors are selected from commercially available options including T-Motor (U8, U10, U12, MN8014), MAD M6C12, and Vector Technics 8020 using the optimization framework described in Section \ref{opt_framework}.

The links adopt a sandwich-style structure~\cite{Stoch3Design}, consisting of a 3D-printed plastic core enclosed between laser-cut aluminum plates. \textbf{Leg link mass} is computed using a parametric model that maps link length to mass. The model estimates the volumes of the aluminum and plastic components and computes the total link mass using their material densities. The resulting mass-length relationship is approximately linear, as shown in Fig.~\ref{fig:mass_v_ll}.

\begin{figure}
\centering
\includegraphics[width=0.7\linewidth]{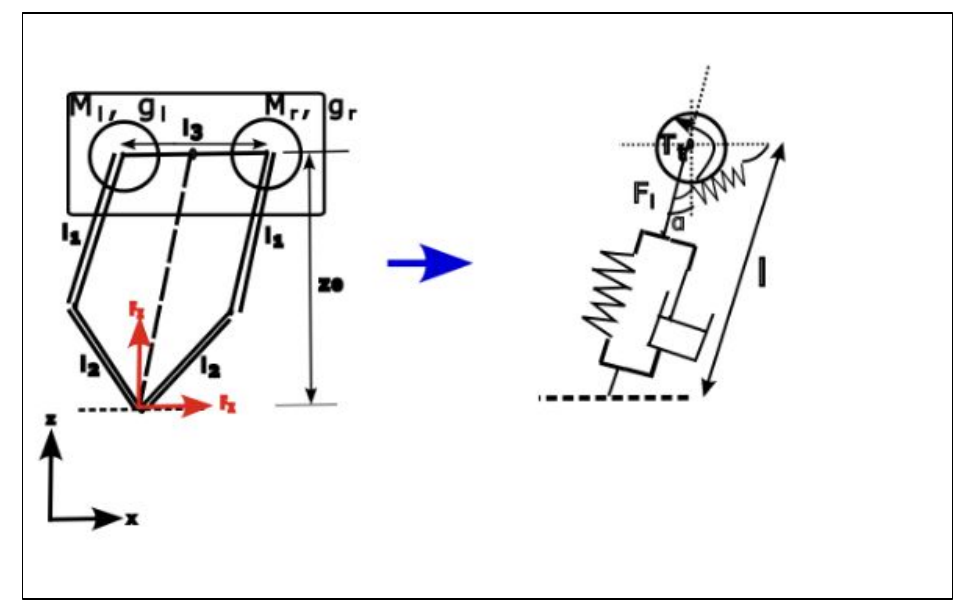}
\caption{The system is modeled as a parallel linear spring--damper of length (l), the distance from the base center to the foot. A torsional spring models angular deflection from a fixed vertical. The robot’s mass is assumed concentrated at base.}
\label{fig:sd}
\vspace{-1.3em}
\end{figure}

\subsection{Monoped Control Architecture}
\label{Controller}

The jumping controller is based on a virtual spring-damper framework inspired by \cite{VMC}. The symmetric five-bar leg is modeled as a virtual system comprising a torsional spring at the revolute joint and a linear spring-damper at the prismatic joint, as shown in Fig.~\ref{fig:sd}.

The virtual spring is assigned a resting length $l_0$. The force generated by the linear spring-damper element is given by: $F_{l}(l, \dot{l}) = K(l_{0} - l) - C\dot{l}$, where $K$ and $C$ denote the linear spring stiffness and damping coefficients, respectively. The torque produced by the torsional spring is: $\tau_{t}(\alpha) = T(\alpha_{0} - \alpha)$, where $\alpha$ is the angular deflection, $\alpha_0$ is the resting orientation, and $T$ is the torsional stiffness constant.

The forces exerted by the virtual spring-damper system on the ground, are resolved along the world
frame $z$- and $x$-axes.The corresponding joint torques are computed as $ \tau = J^{T}F $, where $F = [F_{x}, F_{z}]^{T}$ and $J$ denotes the Jacobian of the five-bar leg. These torques are applied to both actuated hip joints of the planar five-bar mechanism.


\section{Optimization Framework Overview} \label{opt_framework}

\begin{figure}[t]
    \centering
    \includegraphics[width=0.9\linewidth]{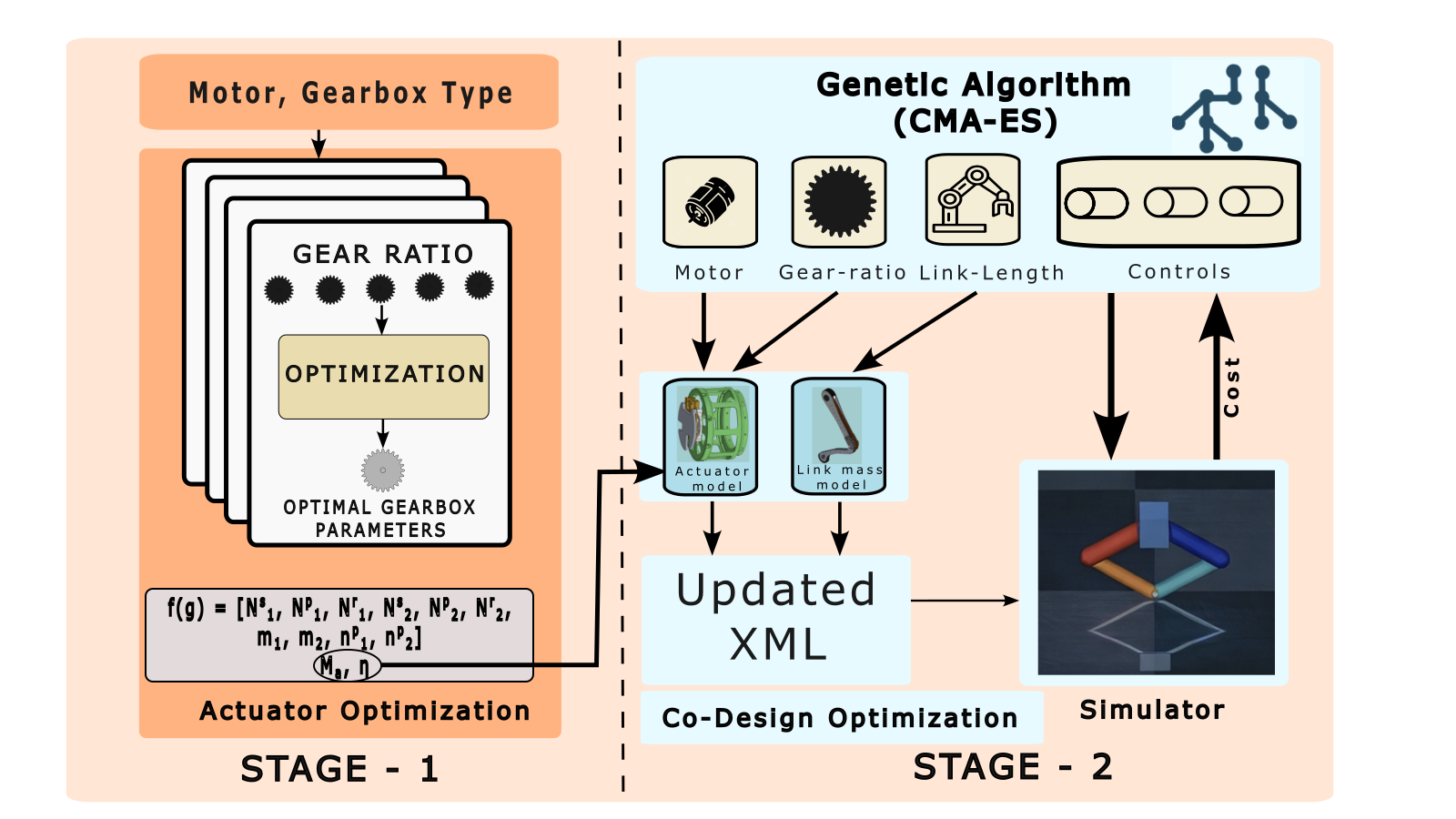}
    \caption{Overview of the co-design framework. \textbf{Stage-1: Actuator Optimization} computes optimal gearbox parameters, mapping gear ratio to actuator mass and efficiency. \textbf{Stage-2: Co-design Optimization} uses CMA-ES to optimize gear ratios, motors, link lengths, and control parameters.}
    \label{fig:main_diag}
    \vspace{-1.3em}
\end{figure}

This section presents a two-stage framework for the co-optimization of monoped design and control parameters to maximize jump distance while minimizing energy consumption.
In Stage~1, a mapping is established between motor selection and gear ratio to corresponding actuator properties, including mass, efficiency, and peak torque, across SSPG, CPG, and WPG gearbox configurations.
In Stage~2, this mapping is utilized to jointly optimize design and control parameters using the CMA-ES algorithm, yielding the optimal configuration for the monoped system. The actuator optimization methodology for both Stage~1 and Stage~2 is illustrated in Fig.~\ref{fig:main_diag}.

\subsection{Stage 1: Actuator Optimization}

\subsubsection{Optimization Variables}
\label{act_vars}
For all three gearbox types (SSPG, CPG, and WPG), the design variables form a column vector 
$X \in \mathcal{X} \subseteq \mathbb{R}^{10}$:
\begin{equation}
    X := [N^s_1, N^p_1, N^r_1, N^s_2, N^p_2, N^r_2,  m_1, m_2, n^p_1, n^p_2],
\end{equation}
where $N^s_i, N^p_i, N^r_i$ are the teeth counts of the sun, planet, and ring gear, 
$m_i$ is the gear module, and $n^p_i$ the number of planets in stage $i$ ($i \in \{1,2\}$). 
All variables are integers except module. Module is chosen from a discrete set.  
As described in Section \ref{five bar arch}, each gearbox type has a different architecture. SSPG has only a single stage.   In both CPG and WPG, as planets are rigidly attached $n^p_1=n^p_2$. Also,  in CPG, $N^r_1=0$.

\subsubsection{Constraints}
\label{act_const}
The optimization includes several constraints which ensure geometric compatibility and manufacturability. These constraints are briefly specified below. More details on each of the constraints are given in \cite{singh2025compact}. 

\begin{enumerate}[label=\Roman*.]

\item \textbf{Gear Ratio Constraint:}
The following constraint ensures the gear ratio of the gearbox remains within the specified range:$GR_{min} \leq GR\leq GR_{max}$. The equations for gear ratios of each type are given in \cite{singh2025compact}. 
\item \textbf{Geometric Constraint:} This constraint ensures the dimensional compatibility of the sun, ring, and planet gears for each stage. It is mathematically represented~\cite{sspgTeethMatchingCond} as:~ $\text{SSPG:}~N^r_i = N^s_i + 2N^p_i;~\text{CPG:}~m_2 N^r_2 = m_1(N^s_1 + N^p_1) + m_2 N^p_2;~\text{WPG:}~ N^r_1 = N^s_1 + 2N^p_1, m_2 N^r_2 = m_1(N^s_1 + N^p_1) + m_2 N^p_2, m_2 N^r_2 < m_1 N^r_1.$
The additional inequality in the WPG is for the stage-1 ring gear to be larger than the stage-2 ring gear, thus enabling higher gear reduction for the same space.
\item \textbf{Meshing Constraints:}  
This ensures proper tooth engagement between gears. These are expressed using the modulo operator, where $a \ \% \ b = 0$ indicates that $a$ is divisible by $b$: $\text{SSPG:}~ (N^s_i + N^r_i) \ \% \ n^p_i = 0;~\text{CPG:}~ N^s_1 \ \% \ n^p_1 = 0, \quad N^r_2 \ \% \ n^p_1 = 0;~\text{WPG:}~(N^s_1 + N^r_1) \ \% \ n^p_1 = 0, N^s_1 \ \% \ n^p_1 = 0, N^r_2 \ \% \ n^p_1 = 0.$


\item \textbf{No Interference Constraints:}  
These constraints prevent collisions between adjacent planet gears and the carrier extrusion. The conditions are formulated using pitch radii~\cite{sspgTeethMatchingCond}: $\text{SSPG:}~
    2(R^s_{i} + R^p_{i}) \sin\!\left(\tfrac{\pi}{2n^p_{i}}\right) - R^p_{i} - R_{ce} \geq \delta_{ce};~\text{CPG/WPG:}~2(R^s_{1} + R^p_{1}) \sin\!\left(\tfrac{\pi}{2n^p_{1}}\right) 
    - R^p_{1} - R_{ce} \geq \delta_{ce}.$
Here, $R^s_i$ and $R^p_i$ denote the pitch radii of the sun and planet gears in stage~$i$, $R_{ce}=4\text{ mm}$ is the carrier extrusion radius, and $\delta_{ce}=1\text{ mm}$ is the minimum clearance to avoid interference. The carrier extrusion connects the front and rear carriers and lies between adjacent planet gears.

\item  \textbf{Additional Constraints:} These constraints limit the range of the optimization variables. They are mathematically represented as: $m_{\min} \leq m \leq m_{\max};~ N_s, N_p \geq N_{\min};~m N_r \leq D^{\max}_{GB};~n^{\min}_p \leq n_p \leq n^{\max}_p.$
\end{enumerate}

These are not applicable to variables that are fixed to zero, for each gearbox type as described in  Section~\ref{act_vars}.In this work, we use $m_{\min} = 0.5$ mm, $m_{\max} = 1.2$ mm, and $N_{\min} = 18$ to avoid undercutting. The number of planet gears is constrained to $n_p \in [2,7]$ based on standard practices.

\subsubsection{Optimization Formulation}  
\label{act_opt_def}

The objective is to minimize actuator mass and maximize efficiency for a given motor and gear ratio. Minimizing actuator mass reduces the overall mass of the robot, which decreases torque requirements and is important for highly dynamic maneuvers such as jumping~\cite{MITActuatorDesign}. Therefore, the optimization framework seeks to minimize actuator mass \(M_{act}\) while maximizing efficiency \(\eta\), subject to the constraints described in Section~\ref{act_const}. A detailed actuator mass model is used to estimate \(M_{act}\). Efficiency equations for \(\eta\) for each gearbox type are given in~\cite{singh2025compact}. For a specific gear ratio \(GR_{req}\), a penalty weighted by \(K_g\) accounts for the deviation between \(GR\) and \(GR_{req}\). The cost function is:
\begin{equation}\label{eq:cost}
    C_{\text{act}} := K_m M_{act} - K_e \eta + K_g |GR_{req} - GR|
\end{equation}
where \(K_m=0.33\), \(K_e=0.67\), and \(K_g=1\) are the weights for mass, efficiency, and ratio tracking, respectively. The optimization problem is:
\begin{equation}
\begin{aligned}
    \min_{\mathcal{X}} \quad  C_{\text{act}},~~~ 
    \text{s.t. Constraints}.
\end{aligned}
\end{equation}

The solution \(C_{act}^*\) gives the optimized value of this objective function for a given gear ratio. The optimization is performed using brute-force search for gear ratios ranging from 4.0:1 to 25.0:1 in increments of 0.1, for each gearbox type. For each gear ratio, the gearbox type that gives the minimum value of the objective function is selected along with its optimized gearbox parameters. This optimization is performed for each motor. Therefore, this framework yields a mapping from gear ratio to actuator mass and efficiency, and in turn peak torque, which is used in Stage~2 of the monoped co-design. The peak actuator torque is computed as the motor peak torque multiplied by the gearbox efficiency and the gear ratio.

\subsection{Stage 2: Co-Design Optimization}
\label{co-opt problem} 
The gear ratio-mass, efficiency and peak torque mapping from Stage 1 is used in Stage 2 to co-optimize mechanical and control parameters. This stage aims to maximize jump distance while considering energy consumption. The methodology is shown in Stage 2 of Fig.~\ref{fig:main_diag}.  
The optimization problem is formulated as follows:


The co-design optimization variables form a column vector \(Y \in \mathcal{Y} \subseteq \mathbb{R}^{13}\):
\begin{equation}\label{co-opt_var}
    Y := [l_{1}, l_{2}, l_{3}, M_{l}, M_{r}, g_{l}, g_{r}, z_{0}, K, C, T, l_{0}, \alpha_{0}]
\end{equation}

Here, \(l_{1}, l_{2}\) are the upper and lower link lengths. Since we optimize a symmetrical five-bar structure, both upper links are equal and both lower links are equal. \(l_{3}\) is the distance between the two actuated joints as shown in Fig.~\ref{fig:sd}. \(g_{r}, g_{l}\) are the gear ratios for the right and left hip joints. \(z_{0}\) is the initial vertical position of the robot base before the controller is applied for a jump (as marked in Fig. \ref{fig:sd}). \(K, C, T, l_{0}, \alpha_{0}\) are control parameters (see Section~\ref{Controller}). All variables are continuous except the motor choice and gear ratios, which are selected from a discrete set. \(\mathcal{Y}\) is the design variable space for this stage.
\subsubsection{Constraints}
\label{sec:co-opt const}
The constraints for this problem are the bounds on all variables: $l_{\min} \leq l_{1}, l_{2} \leq l_{\max};~      a_{\min} \leq l_{3} \leq a_{\max};~
    g_{\min} \leq g_{r}, g_{l} \leq g_{\max};~
    K_{\min} \leq K \leq K_{\max};~
    C_{\min} \leq C \leq C_{\max};~
    T_{\min} \leq T \leq T_{\max};~
    b_{\min} \leq l_{0} \leq b_{\max};~
    \alpha_{\min} \leq \alpha_{0} \leq \alpha_{\max};~ 
    0 \leq z_{0} \leq l_{1} + l_{2};~
    M_{l}, M_{r}\in \{1,2,3,4,5,6\}.$

We use \(l_{\min} = 0.15\) m, \(l_{\max} = 0.35\) m, \(a_{\min} = 0.05\), \(a_{\max} = 0.15\), \(g_{\min} = 4\), and \(g_{\max} = 25\). Control variable limits are \(K_{\min} = 50\), \(K_{\max} = 1000\); \(C_{\min} = 0\), \(C_{\max} = 10\); and \(T_{\min} = 10\), \(T_{\max} = 50\). The lower and upper bounds are heuristic. For \(K\) and \(T\), increasing the limits beyond the upper bounds does not improve maximum distance due to actuator torque limits set by the selected motors and gear ratios. Additionally, \(b_{\min} = 0.2\) m, \(b_{\max} = 7\) m, \(\alpha_{\min} = -\pi/2\), and \(\alpha_{\max} = \pi/2\). \(z_{0}\) is limited by the sum of link lengths, as higher values are not reachable by the robot when in contact with the ground. The motors are selected from a fixed set: (1) T-Motor U8, (2) T-Motor U10, (3) T-Motor U12, (4) T-Motor MN8014, (5) Vector Technics VT8020, and (6) MAD M6C12.

\subsubsection{Optimization formulation} 

The objective is to maximize jump distance while minimizing mechanical energy consumed. Greater jump distance results in better terrain gap coverage. The Stage~2 cost is defined as  
\begin{equation}
\label{comb cost}
    C_{c}(\eta, \tau, \omega) = \lambda_{1} C_{d}(\eta,\tau) + \lambda_{2} C_{e}(\eta, \tau, \omega)
\end{equation}
where \(C_{d}\) is the jump distance cost, \(C_{e}\) is the mechanical energy consumed for one jump, and \(\lambda_{1}, \lambda_{2}\) are their respective weights. The calculations of \(C_{d}\) and \(C_{e}\) are detailed below:

\begin{equation}
\label{distance cost}
    C_{d}(\eta, \tau) = -K_{d} {|x(\eta, \tau)|}
\end{equation}
where \(x\) is the jump distance along the x-axis attained by the robot at the end of a jump. It is measured as the position of the robot base along the x-axis at the end of the jump with respect to its start position. \(K_{d}\) is a scaling constant chosen to keep \(C_{d}\) and \(C_{e}\) of similar order in~\eqref{comb cost}. The variable \(\tau = \begin{bmatrix} \tau_l & \tau_r \end{bmatrix} \in \mathbb{R}^{m \times 2}\) contains torque sequences generated by the controller (see Section~\ref{Controller}) for the left hip (\(\tau_l \in \mathbb{R}^m\)) and right hip (\(\tau_r \in \mathbb{R}^m\)) joints over one jump, where \(m\) is the number of simulation timesteps in a single jump. The actual torques applied by the actuators are multiplied by the actuator efficiency values obtained from Stage~1 (\(\eta_{l}\) for the left actuator and \(\eta_{r}\) for the right actuator). \(x\) depends only on torques applied until the foot is in stance phase or in contact with the ground. Control parameters \(K\), \(C\), \(T\), \(l_{0}\), and \(\alpha_{0}\) directly influence \(\tau\), as shown in Section~\ref{Controller}. 

The mechanical energy consumed during a jump is computed by summing the energies consumed by the left and right hip actuators over all timesteps. For actuator \(j \in \{l,r\}\) at timestep \(i\),
\begin{equation}
E_{j}(\eta_{j},\tau_{j}(i),\omega_{j}(i)) =
\begin{cases}
\eta_{j}\tau_j(i) \, \omega_j(i) \, dt, & \tau_j(i) \, \omega_j(i) > 0, \\
0, & \text{otherwise}.
\end{cases}
\end{equation}
This excludes regenerative energy as only motor power applied to the system is considered. The total energy is
\begin{equation}
C_e(\eta,\tau,\omega) = \sum_{i=1}^{m} \big( E_l(\eta_{l},\tau_{l}(i), \omega_{l}(i))
+ E_r(\eta_{r},\tau_{r}(i), \omega_{r}(i)) \big)
\end{equation}
where \(\omega_l, \omega_r\) are angular velocities at the left and right hip, obtained from the simulator. In matrix form,
\[
\tau = \begin{bmatrix} \tau_l & \tau_r \end{bmatrix}, 
\omega = \begin{bmatrix} \omega_l & \omega_r \end{bmatrix} \in \mathbb{R}^{m \times 2},
\eta = \begin{bmatrix} \eta_l & \eta_r \end{bmatrix} \in \mathbb{R}^{1 \times 2}.
\]
The Stage~2 optimization problem is:
\begin{equation}
\begin{aligned}
    \min_{\mathcal{Y}} ~~\lambda_{1}C_{d}(\eta,\tau) + \lambda_{2}{C_{e}(\eta, \tau, \omega)},~\text{s.t.}~\text{Constraints (\ref{sec:co-opt const})}
\end{aligned}
\end{equation}

We solve the Stage~2 optimization problem using the Covariance Matrix Adaptation Evolution Strategy (CMA-ES)~\cite{CMAES}, which samples \(N\) candidate solutions from a multivariate normal distribution with an initial mean and step-size \(\sigma\) controlling the search radius. Each sample defines design and control parameters, with design variables generating XML files for the MuJoCo simulator~\cite{mujoco}. Gear ratios and motors selected in the samples generated by CMA-ES update actuator mass, efficiency, and peak torque limits via the mapping obtained from Stage~1, while link lengths in the samples update monoped link dimensions and masses using the leg-link mass model (Section~\ref{five bar arch}). The simulator evaluates each sample on a single-jump task, computes a cost, and CMA-ES updates the distribution mean \(\mu\) and step-size \(\sigma\) based on current-generation costs.

\begin{table*}[!b]
\centering
\caption{\small{Optimized Design and Control Parameters for Different Cases}}
\label{tab:design_params}
\resizebox{1.0\textwidth}{!}{\begin{tabular}{|l|c|c|c|c|c|c|c|c|c|c|c|c|c|c|c|}
\hline
\textbf{Case} & \textbf{$\mathbf{l_1 (m)}$} & \textbf{$\mathbf{l_2 (m)}$} & \textbf{$\mathbf{l_3 (m)}$} & \textbf{$\mathbf{z_{0} (m)}$} & \textbf{$\mathbf{M_{l}}$} & \textbf{$\mathbf{M_{r}}$} & \textbf{$\mathbf{g_{l}}$} & \textbf{$\mathbf{g_{r}}$} & \textbf{$\mathbf{K}$} & \textbf{$\mathbf{C}$} & \textbf{$\mathbf{T}$} &
\textbf{$\mathbf{l_{0}}$} &
\textbf{$\mathbf{\alpha_{0}}$} &
\textbf{$\mathbf{X(m)}$} & \textbf{$\mathbf{E(J)}$}\\
\hline
A      & 0.19 & 0.3 & 0.05 & 0.3  & U10 & U10  & 6.0:1 & 6.0:1 & 926.8 & 1.7 & 35.7 & 0.2  & 0.62 & 0.58 & 13.18 \\

B      & 0.297 & 0.302 & 0.1 & 0.424 & U8 & MAD-M6C12 & 4.0:1 & 21.4:1 & 916.2 & 6.9 & 10.2 & 0.22 & 1.03 & 0.97 & 22.72 \\

\textbf{C} 
       & \textbf{0.207} & \textbf{0.322} & \textbf{0.076} & \textbf{0.32} 
       & \textbf{U8} & \textbf{MN8014} 
       & \textbf{16.2:1} & \textbf{4:1} 
       & \textbf{1000} & \textbf{2.0} & \textbf{32.2} 
       & \textbf{0.2} & \textbf{-1.18} & \textbf{1.03} & \textbf{23.0} \\

Nominal  & 0.297 & 0.302  & 0.1 & 0.3 & U10  & U10 & 6.0:1 & 6.0:1 & 670 & 5.7 & 10.1 & 0.22 & 1.075 & 0.79 & 26.0 \\
\hline
\end{tabular}
}
\end{table*}

\begin{table*}[!b]
\centering
\caption{\small{Optimized Actuator Parameters of optimal motor and gear ratios from Stage 2}}
\label{tab:actuator_designs}
\resizebox{1.0\textwidth}{!}{\begin{tabular}{|l|c|c|c|c|c|c|c|c|c|c|c|c|}
\hline
\textbf{Case}& \textbf{$\mathbf{M_{l}}$} & \textbf{$\mathbf{g_{l}}$} & \textbf{Type} & \textbf{mass} & \textbf{$\eta$} &
\textbf{Parameters} & \textbf{$\mathbf{M_{r}}$} & \textbf{$\mathbf{g_{r}}$} & \textbf{Type} & \textbf{mass} & \textbf{$\eta$} &
\textbf{Parameters} \\
 & &  & & & & $[N^s_i, N^p_i, N^r_i,  n^p_i, m_i]$ $i \in \{1,2\}$ & &  & & & & $[N^s_i, N^p_i, N^r_i,  n^p_i, m_i]$ $i \in \{1,2\}$  \\
\hline
A &  U10 & 6:1 & SSPG & 0.853  & 0.952 & \textbf{Stage 1:}$[22 , 44 , 110, 3, 0.6]$  & U10 & 6:1 & SSPG & 0.853 & 0.952 & \textbf{Stage 1:}$[22 , 44 , 110 , 3, 0.6]$
 \\
 & & & & & &  \textbf{Stage 2:}$[0, 0, 0, 0, 0]$ & & & & & & \textbf{Stage 2:}$[0, 0, 0, 0, 0]$            
\\
\hline
B & U8 & 4:1 & CPG & 0.523 & 0.963 & \textbf{Stage 1:}$[30 ,40 ,0,3, 0.5]$  & MAD-M6C12 & 21.4:1 & WPG & 0.769 & 0.871 & \textbf{Stage 1:}$[28 ,46 ,120 ,4, 0.6]$ 
 \\
 & & & & & &  \textbf{Stage 2:}$[0,56, 126 ,3, 0.5]$ & & & & & & \textbf{Stage 2:}$[0, 30, 104, 4, 0.6]$ 
 \\
\hline
\textbf{C} & \textbf{U8} & \textbf{16.2:1} & \textbf{WPG} & \textbf{0.713} & \textbf{0.893} & \textbf{Stage 1:}\textbf{$[32, 48, 128, 4, 0.5]$}  & \textbf{MN8104} & \textbf{4:1} & \textbf{SSPG} & \textbf{0.739} & \textbf{0.962} & \textbf{Stage 1:}\textbf{$[39 ,39 ,117 ,3, 0.5]$}
 \\
 & & & & & &  \textbf{Stage 2:}$[0, 28, 108, 4, 0.5]$ & & & & & & \textbf{Stage 2:}$[0, 0, 0, 0, 0]$
 \\
\hline
Nominal &  U10 & 6:1 & SSPG & 0.853  & 0.952 & \textbf{Stage 1:}$[22 , 44 , 110, 3, 0.6]$  & U10 & 6:1 & SSPG & 0.853 & 0.952 & \textbf{Stage 1:}$[22 , 44 , 110 , 3, 0.6]$
\\
& & & & & &  \textbf{Stage 2:}$[0, 0, 0, 0, 0]$ & & & & & & \textbf{Stage 2:}$[0, 0, 0, 0, 0]$ 
\\
\hline
\end{tabular}
}
\label{table:Opt_param}

\end{table*}

\section{Results and Discussion}
\label{results}
\subsection{Actuator Optimization Results}
\label{Act_results}
\begin{figure}[t]
    \centering
    \includegraphics[width=0.6\linewidth]{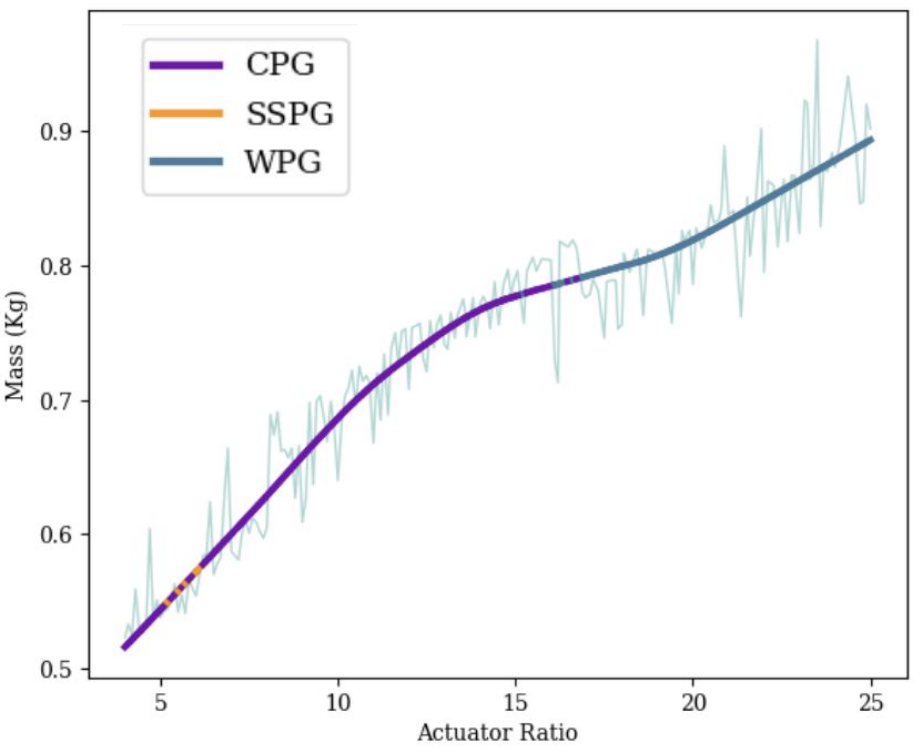}
    \caption{Optimized actuator mass vs gear ratio for U8 motor}
    \label{fig:mass vs gr}
    \vspace{-1em}
\end{figure}

\begin{figure}[t]
    \centering
    \includegraphics[width=0.62\linewidth]{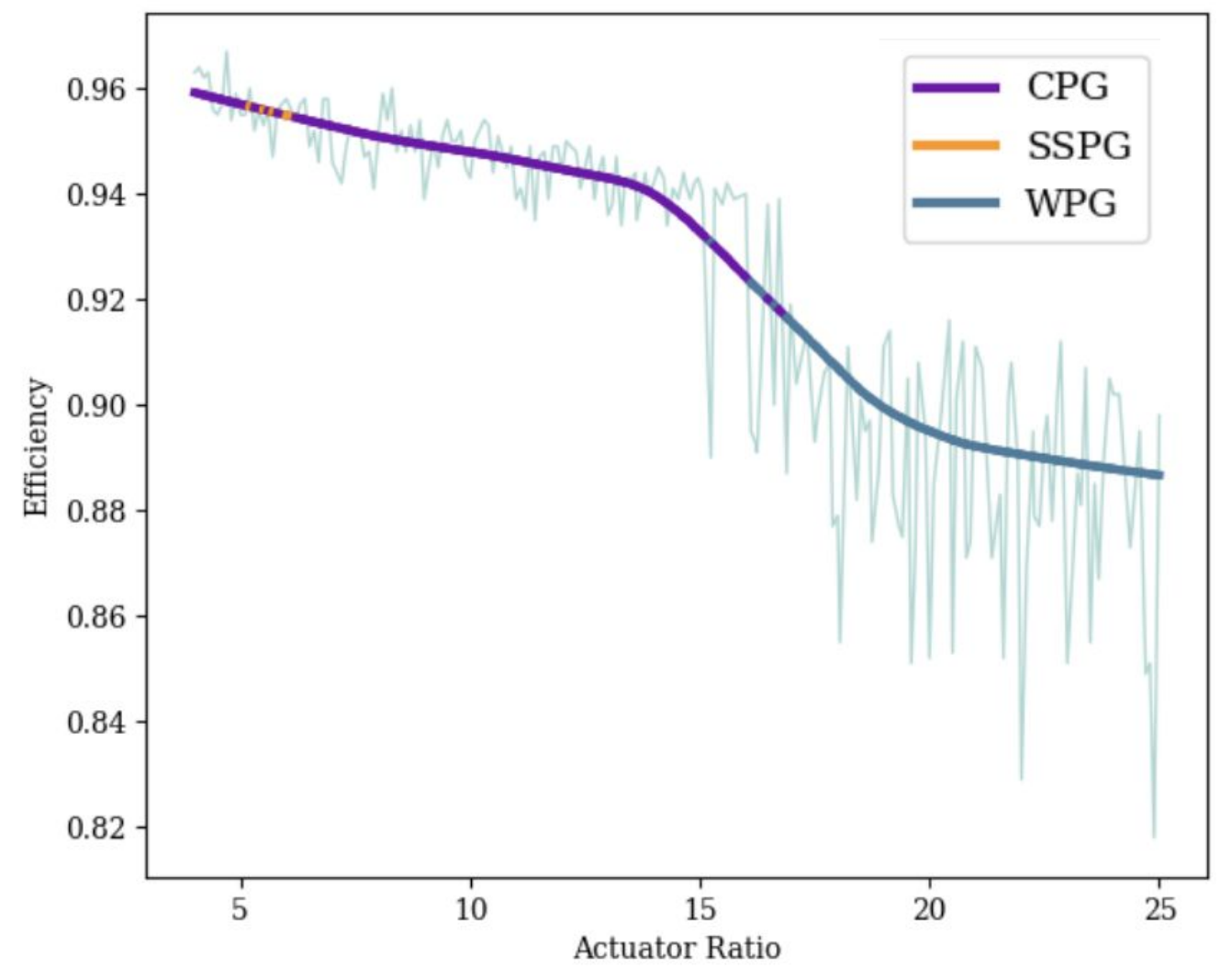}
    \caption{Optimized efficiency vs gear ratio for U8 motor}
    \label{fig:eff vs gr}
    \vspace{-1em}
\end{figure}

This subsection presents the Stage 1 actuator optimization results over gear ratios ranging from $4.0{:}1$ to $25.0{:}1$. For each motor and gearbox type, a mapping is generated between gear ratio and the corresponding optimal gearbox parameters, including teeth count, module, and number of planets. At each discrete gear ratio (0.1 increments), the configuration minimizing (\ref{eq:cost}) is selected. The procedure is repeated for all motors described in Section~\ref{act_const}. Actuator mass, comprising components dependent on motor size and gearbox parameters, is modeled as a function of gear ratio and motor properties, while efficiency is computed following \cite{singh2025compact}. The optimization is performed for SSPG, CPG, and WPG gearbox types. Figures~\ref{fig:mass vs gr} and \ref{fig:eff vs gr} show the resulting mass and efficiency trends for the T-Motor U8. The following observations can be made:
\begin{enumerate}
\item The feasible design space varies across gearbox types: SSPG is limited to ratios up to $7.2{:}1$, CPG up to $17.9{:}1$, and WPG spans the $10{:}1$--$25{:}1$ range.
\item For the T-Motor U8, CPG provides the lowest cost over most of the $4{:}1$--$16.8{:}1$ range, while WPG becomes optimal for higher ratios ($16.9{:}1$--$25{:}1$).
\item Actuator mass increases with gear ratio for both CPG and WPG configurations (Fig.~\ref{fig:mass vs gr}). Efficiency generally decreases with increasing gear ratio, while remaining above $90\%$ up to approximately $16{:}1$ (Fig.~\ref{fig:eff vs gr}).
\end{enumerate}

\subsection{Co-Design Optimization Results}
\label{co-opt results}
This subsection presents the results of co-optimizing the design and control parameters of a planar five-bar monoped to maximize hop distance while minimizing energy consumption, and compares them against a nominal design. The nominal configuration (Table~\ref{tab:design_params}) consists of a symmetrical five-bar mechanism with link lengths ($l_{1}, l_{2}$) and identical motor-gearbox combinations at both actuated hip joints. The link lengths, motors, and gear ratios are similar to \cite{Stoch3Design}. The corresponding actuator parameters are obtained using the Stage 1 optimization framework (Section~\ref{act_opt_def}), and the optimized controller parameters and performance metrics are reported in Table ~\ref{tab:design_params}. Two ablation studies are additionally performed to evaluate the effects of (i) link length optimization and (ii) actuator optimization. Three cases are considered: Case-A optimizes link lengths ($l_{1}, l_{2}, l_{3}$) and control parameters while keeping actuators fixed; Case-B optimizes motor selection, gear ratios, and control parameters with fixed link lengths; and Case-C jointly optimizes all design and control parameters ($l_{1}, l_{2}, l_{3}, M_{l}, M_{r}, g_{l}, g_{r}, z_{0}, K, C, T, l_{0}, \alpha_{0}$). All designs are generated using CMA-ES. Simulations are performed in MuJoCo with a timestep of 0.001~s, with control torques applied only during ground contact. The gear ratio to mass and efficiency mappings are obtained from the actuator optimization stage (Section~\ref{Act_results}). 



\subsubsection{Case A}

To evaluate the influence of link lengths ($l_1$, $l_2$) and the distance between actuated hip joints ($l_3$), the actuators were kept identical to the nominal configuration while optimizing $l_{1}, l_{2}, l_{3}$ and the control parameters. This configuration achieves the lowest energy consumption among all cases, reducing mechanical energy by approximately $49.3\%$ ($26.0$ J to $13.18$ J), but with reduced jump distance ($0.79$ m to $0.58$ m). The optimized design and actuator parameters are summarized in Tables~\ref{tab:design_params} and~\ref{table:Opt_param}, respectively.

\subsubsection{Case B}

To evaluate the effect of actuator selection and transmission ratios, the link dimensions were fixed to the nominal configuration while optimizing motors, gear ratios, and control parameters. The resulting design employs asymmetric transmission characteristics, with a higher gear ratio at the right actuator ($g_{r}=21.4{:}1$) and a lower ratio at the left actuator ($g_{l}=4{:}1$). This configuration improves jump distance by approximately $22.7\%$ (from $0.79$ m to $0.97$ m) while reducing energy consumption by about $12.6\%$ (from $26.0$ J to $22.72$ J). The optimized actuator configuration consists of a WPG gearbox for the right actuator and a CPG gearbox for the left actuator, as summarized in Table~\ref{table:Opt_param}.

\subsubsection{Case C}
Case C corresponds to the fully co-optimized configuration, where both link dimensions and actuator parameters are optimized simultaneously. Compared to the nominal case, this configuration achieves the highest performance improvement, increasing jump distance by approximately $30.4\%$ (from $0.79$ m to $1.03$ m) while reducing energy consumption by about $11.5\%$ (from $26.0$ J to $23.0$ J). The optimized actuator configuration consists of a WPG gearbox for the left actuator and an SSPG gearbox for the right actuator, as detailed in Table~\ref{table:Opt_param}. Despite requiring energy comparable to Case B, this configuration achieves the largest increase in jump distance.

\begin{figure}
    \centering
    \includegraphics[width=0.9\linewidth]{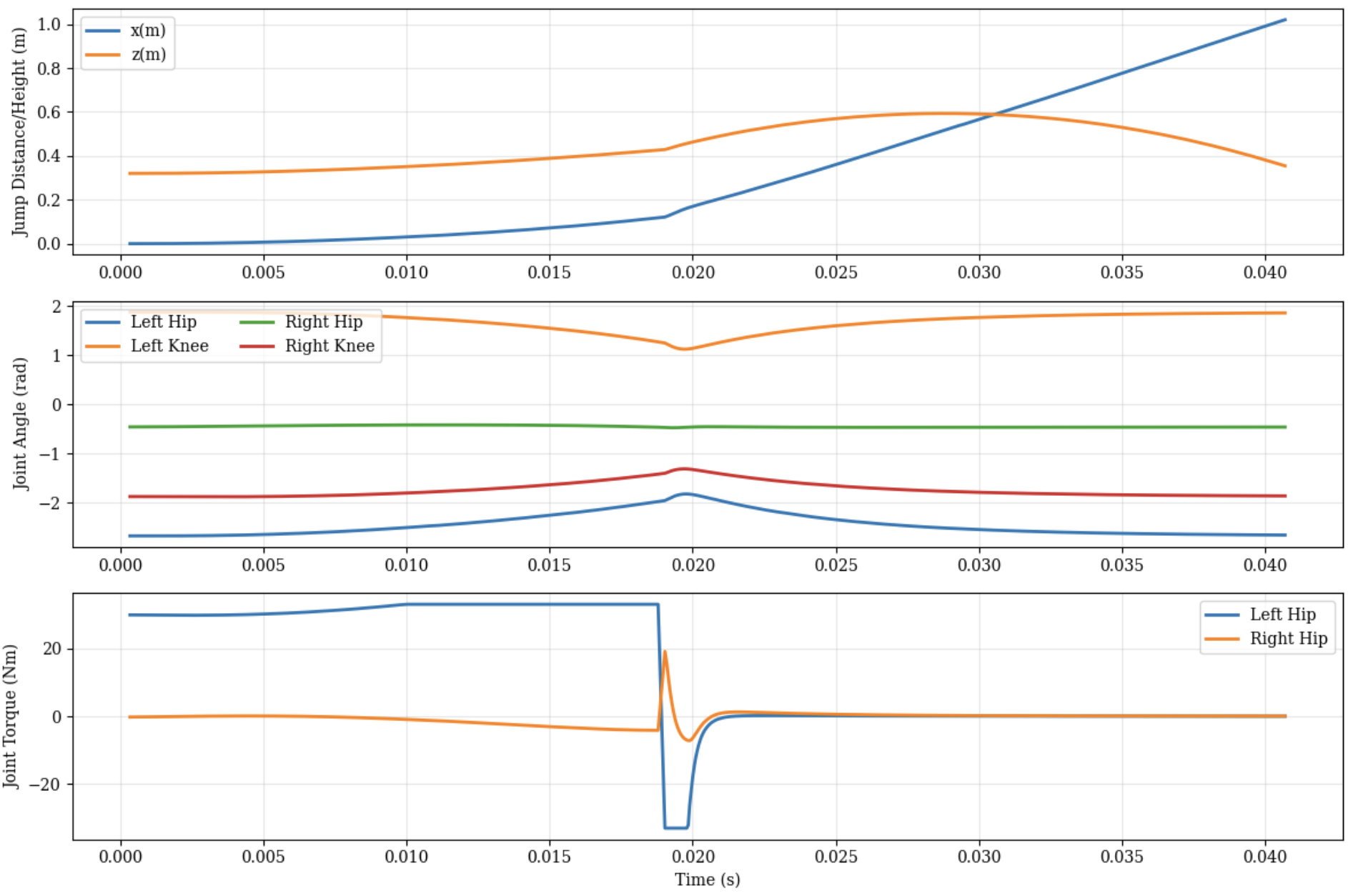}
    \caption{Optimal trajectories, Case C}
    \label{fig:placeholder}
    \vspace{-1.3em}
\end{figure}

Overall, Case A shows that optimizing link lengths alone significantly reduces energy consumption but degrades jump distance. In contrast, Case B demonstrates that actuator optimization enables simultaneous improvement in jump distance and energy consumption, indicating a stronger influence of motor and gear ratio selection on performance. Case C achieves the best trade-off, further increasing jump distance while maintaining comparable energy consumption, highlighting the benefits of jointly optimizing morphology, actuation, and control parameters.

\vspace{-0.4em}
\section{Conclusion}
\label{conclusion}
\vspace{-1mm}
This paper presented a novel co-design framework for a planar five-bar monoped that jointly optimizes mechanical design, actuator parameters, and control for dynamic jumping. By incorporating actuator-level design through a gear-ratio-dependent mapping of mass, efficiency, and torque, the proposed approach achieved a 30.4\% increase in jump distance and an 11.5\% reduction in mechanical energy consumption compared to a nominal design. These results demonstrate the value of integrating detailed actuator modeling within co-design frameworks for closed-loop mechanisms. Future work will focus on extension to multi-joint legged robots.

\bibliographystyle{IEEEtran}
\bibliography{references}

\end{document}